\documentclass[conference]{IEEEtran}
\IEEEoverridecommandlockouts
\usepackage{amsmath,amsfonts} 
\usepackage{graphicx, caption, subcaption}
\usepackage{algorithmic}
\usepackage{float}
\usepackage{graphicx}
\usepackage{textcomp}
\usepackage{tabularray}
\usepackage{multirow}
\usepackage{graphicx}
\usepackage{booktabs}
\usepackage{float}
\usepackage{url}
\usepackage[numbers]{natbib} 
\usepackage{xcolor}
\usepackage[ruled,vlined]{algorithm2e}
\usepackage{caption}
\usepackage{enumitem}

\def\BibTeX{{\rm B\kern-.05em{\sc i\kern-.025em b}\kern-.08em
    T\kern-.1667em\lower.7ex\hbox{E}\kern-.125emX}}
\begin{document}

\newcommand{\name}{{\em {$\mu$}DAR} }
\newcommand{\names}{{\em {$\mu$}DAR}}
\newcommand{\namet}{{$\mu$}DAR's}
\newcommand{\nameb}{{$\mu$}DAR}

\title{Unsupervised Domain Adaptation for Action Recognition via Self-Ensembling and Conditional Embedding Alignment}
\author{\IEEEauthorblockN{\textbf{\author{
  Indrajeet Ghosh$^1$, Garvit Chugh$^2$, Abu Zaher Md Faridee$^3$, Nirmalya Roy$^1$\\
  $^1$Department of Information Systems, University of Maryland, Baltimore County, United States \\ $^2$Department of Computer Science and Engineering, Indian Institute of Technology Jodhpur, India \\
  $^3$Amazon, United States\\
  \texttt{{$^1$(indrajeetghosh, nroy)@umbc.edu,} {$^2$chugh.2@iitj.ac.in,} {$^3$abufari@amazon.com}} \\}}}}
\maketitle
\begin{abstract}

Recent advancements in deep learning-based wearable human action recognition (wHAR) have improved the capture and classification of complex motions, but adoption remains limited due to the lack of expert annotations and domain discrepancies from user variations. Limited annotations hinder the model's ability to generalize to out-of-distribution samples. While data augmentation can improve generalizability, unsupervised augmentation techniques must be applied carefully to avoid introducing noise. Unsupervised domain adaptation (UDA) addresses domain discrepancies by aligning conditional distributions with labeled target samples, but vanilla pseudo-labeling can lead to error propagation. To address these challenges, we propose \textit{$\mu$DAR}, a novel joint optimization architecture comprised of three functions: (i) consistency regularizer between augmented samples to improve model classification generalizability, (ii) temporal ensemble for robust pseudo-label generation and (iii) conditional distribution alignment to improve domain generalizability. The temporal ensemble works by aggregating predictions from past epochs to smooth out noisy pseudo-label predictions, which are then used in the conditional distribution alignment module to minimize kernel-based class-wise conditional maximum mean discrepancy~($k$CMMD) between the source and target feature space to learn a domain invariant embedding. The consistency-regularized augmentations ensure that multiple augmentations of the same sample share the same labels; this results in (a) strong generalization with limited source domain samples and (b) consistent pseudo-label generation in target samples. The novel integration of these three modules in \textit{$\mu$DAR} results in a range of~\textbf{\textit{$\approx$ 4-12\%}} average macro-F1 score improvement over six state-of-the-art UDA methods in four benchmark wHAR datasets. 


\end{abstract}

\vspace{-0.3cm}

\footnotetext[1]{This work has been partially supported by NSF CAREER Award \#1750936, ONR Grant \#N00014-23-1-2119, U.S. Army Grant \#W911NF2120076 and NSF CNS EAGER Grant \#2233879.}
\footnotetext[2]{This work has been accepted to the Proceedings of the IEEE ICDM 2024.}
\section{Introduction}

Over the past decade, wearable human activity recognition (wHAR) has enabled applications in healthcare~\cite{ramamurthy2021star}, sports analytics~\cite{ghosh2022permtl}, and fitness tracking~\cite{xie2018evaluating}, leveraging inertial, gyroscopic, and magnetometric sensors in smartphones and smartwatches to capture motion data, with neural processing units facilitating real-time inference. However, traditional models rely on supervised learning and extensive labeled data, limiting scalable, robust activity recognition. While Activities of Daily Living (ADLs) datasets cover simple actions like \textit{running} and \textit{walking}, micro-complex activities show user-specific patterns, leading to cross-user variations~\cite{sun2011new} due to differences in body characteristics and skill levels~\cite{ghosh2022decoach}. These variations are particularly evident in skill-based actions, where proficiency evolves; for example, experienced badminton players exhibit more consistent patterns than novices~\cite{ghosh2022permtl}, reflecting how skill progression influences activity patterns.

In this context, unsupervised domain adaptation (UDA)~\cite{liu2022deep} offers a viable solution, particularly when annotations are limited and domain discrepancies arise from out-of-distribution samples. UDA leverages labeled data from the source domain (e.g., female, expert player) and unlabeled data from the target domain (e.g., male, novice player), aligning feature distributions to learn a domain-invariant feature space. Most UDA research~\cite{barbosa2018unsupervised} focuses on marginal feature alignment, with few models addressing conditional distribution alignment, often requiring labeled target data~\cite{tachet2020domain}. Pseudo-labeling can be an alternative, but its self-labeling nature often amplifies initial classification errors~\cite{stikic2008exploring}, hindering effective conditional alignment across domains.

To address these challenges, we introduce a novel joint optimization framework using complementary loss components. We begin by training a deep learning classifier on labeled source data. Next, we adapt this model to unlabeled target data by learning a domain-invariant embedding via minimizing the class-wise kernel-based conditional maximum mean discrepancy (kCMMD) distance, using pseudo-labels from the target classifier. To prevent training collapse due to incorrect pseudo-labels, we employ a temporal ensemble approach to smooth them via aggregation. To enhance generalizability, we apply unsupervised data augmentation to both source and target data, coupled with consistency training, by minimizing the KL divergence between predictions of original and augmented samples. This ensures that augmented samples receive consistent labels, increasing robustness to data noise and covering uncertain regions in the decision space. It also maintains consistent pseudo-labels for target samples across augmentations, which, along with the temporal ensemble, provides high-quality pseudo-labels for conditional kernel regularization. This synergy minimizes the domain gap and improves target domain classification without labeled samples. Consequently,~\name outperforms state-of-the-art~(SOTA) UDA baselines on four activity recognition datasets, covering both simple daily activities and complex sports/gym actions.

The key contributions are summarized as follows:
\begin{itemize}[leftmargin=*]
    \item We propose a novel UDA framework (\name) to address two challenges in wHAR: (i) \textit{cross-user variations} and (ii) \textit{limited labeled data}. Our method learns source-specific representations from labeled data and applies them to unlabeled target data. We open-source~{\bf $\mu$\textit{DAR}}{\footnote{\url{https://github.com/indrajeetghosh/uDAR_ICDM}}} for reproducibility.

    \item We develop a novel joint optimization architecture incorporating: (i) \textit{kernel-based class-wise conditional maximum mean discrepancy} ($k$CMMD) to align conditional distributions between source and target features; (ii) a \textit{temporal ensemble-based pseudo-labeling} technique for robust target labels; and (iii) a \textit{consistency regularizer between augmentations and real samples} to enhance generalizability and robustness, improving pseudo-labels of target domain. We detail the synergy of these components via ablation studies.

    \item We compare \textit{$\mu$DAR's} macro-F1 performance against six SOTA UDA algorithms~\cite{hu2023swl,sanabria2021contrasgan,eldele2022cotmix,liu2021adversarial,chen2020homm,mirza2022norm} on four public benchmark datasets~\cite{ghosh2022decoach,altun2010human,reiss2012creating,sharshar2022mm} covering varying label complexities, dexterity levels (novice vs.\ expert). \name{} consistently outperforms baselines by \textbf{$\approx$  4--12\%} over five independent trials per dataset.

    \end{itemize} 

\vspace{-0.3cm}
\section{Related Work}\label{sec:related}

We discuss the relevant literature to \name on UDA in the wHAR and consistency regularization area with an emphasis on handling cross-user variations and limited labeled data and to differentiate \name from the SOTA approaches.

\subsubsection{Wearable-based Unsupervised Domain Adaptation~(UDA)} have been significant advances in wearable-based recognition of daily human activities~\cite{khalifa2017harke}, scaling these models to more niche areas where annotations are scarce has been an open challenge, especially for activities that require expert execution like sports and gym activities. UDA has become a valuable method for overcoming the challenge of limited labelled data as it capitalizes on the abundance of available unlabeled data to enhance the model's generalizability and robustness. One of the few works incorporating UDA to tackle limited label data is \cite{hu2023swl}, where the authors introduced a novel sample differentiation technique, utilizing a parameterized network to identify whether a sample is from the source or target domain and additionally involves assigning weights to pseudo labels for target samples based on the confidence level of the domain classifier. However, recent SOTA studies have addressed minimal label information by minimizing conditional feature distributions. In contrast, this work focuses on tackling limited label data by minimizing conditional discrepancies.

\subsubsection{Consistency Regularization} has recently become prominent, especially in developing robust models invariant to data variations, i.e., augmented samples~\cite{yang2023sample}. The core principle of consistency regularization is to maintain consistent predictions between real and augmented samples. This approach promotes smoothness and low-density separation, crucial for semi-supervised learning environments, aiding in minimizing consistency loss~\cite{englesson2021consistency}. Studies~\cite{xie2020unsupervised} indicate that combining consistency regularization with unsupervised data augmentation in semi-supervised learning impedes performance degradation and bolsters model generalizability and robustness. Motivating by this, we integrate a consistent regularization approach to tackle user variations. However, to design real-world settings, we incorporate unsupervised data augmentation for wHAR~\cite{um2017data}, and to our best knowledge, our work is among the first few that have integrated unsupervised data augmentation with consistency regularization in wHAR. 


\begin{figure}
    \centering
    \includegraphics[scale = 0.26]{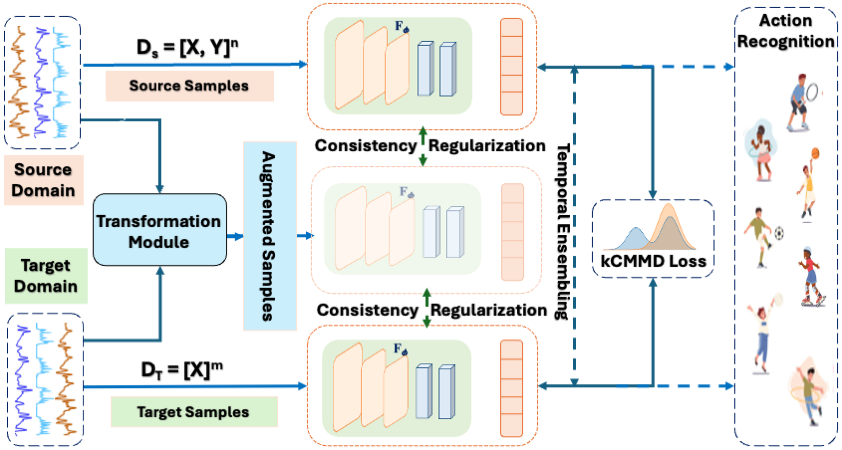}
    \caption{High level overview of the \name architecture}
    
    \label{fig:overview}
\end{figure}

\section{Methodology}
\label{sec:methodology}

In this section, we present the \name framework for the wHAR domain, designed to recognize ADL actions, including gym and sports activities. We briefly outline the problem formulation and highlight the key functional and learning components of \textit{$\mu$DAR}.


\subsection{Problem Formulation}

Cross-user variations and limited expert-labeled datasets hinder the effectiveness of wHAR methods. We leverage labeled data from experienced users (source domain), $D_s = \{(x_i, y_i)\}_{i=1}^{n}$, and unlabeled data from new users (target domain), $D_t = \{x_j\}_{j=1}^{m}$. Here, $n$ and $m$ represent the number of instances in the source and target domains, respectively, with both sharing the same label space, $y_s^i, {y_t^j} \in \{1, 2, \ldots, c\}$, where $c$ denotes the number of activity classes. UDA aims to identify activities for new users by utilizing labeled source data alongside unlabeled target data. This assumes that both domains use identical sensors and placements, performing the same activities, despite conditional distribution shifts, i.e., $P(y\,|\,x_s) \neq P(y\,|\,x_t)$.


Figure~\ref{fig:overview} illustrates the proposed UDA framework, comprising four key components to tackle these challenges: (i) learning source-specific features to capture trained user characteristics; (ii) generating high-quality pseudo labels via temporal ensembling; (iii) aligning source and target features using the proposed \(k\)CMMD loss; and (iv) applying unsupervised data augmentation to improve robustness against user-variations. This integrated approach minimizes domain discrepancies, enhancing generalization to new users in the wHAR domain.

\subsection{Supervised Learning Source-specific Feature Learning} 

We explore the supervised training process designed to capture the unique features of the source domain data. Our approach involves a specifically tailored supervised training architecture. This architecture consists of three convolutional layers, each with an increasing number of filters and progressively smaller kernel sizes. Each convolutional layer is coupled with batch normalization and max-pooling layers to minimize internal covariate shifts and enhance learning of the high and low-level features, respectively. Followed by two fully connected layers and a softmax layer for output. The training objective is to minimize the \textit{\textbf{CE supervised loss}} {$\mathcal{L}_{SL} = -\frac{1}{n} \sum_{i=1}^{n} P(y_{s_i} | x_{s_i}) \log P(\hat{y}_{s_i} | x_{s_i})$\} and ensure accurately predict the class distribution $P(\hat{Y}|X)$. 


\subsection{Extracting Pseudo Labels using Temporal Ensembling}

Adapting our supervised training pipeline from a well-labeled source domain to a target domain devoid of labeled data presents a significant challenge. This is further complicated by activities' inherent variability and complexity, especially when dealing with data from synchronized sensors on multiple wearable devices. To effectively generate high-quality pseudo labels in this context, we implement \textit{temporal ensembling}~\cite{laine2016temporal}. This method is designed to maximize the concurrent extraction of mutual information from both the source and target domains, which aims to enhance the pseudo-label generation process by aggregating model predictions in an iterative refinement update method highlighted in Eq.~\ref{eq:weight_update}, where $\mathcal{L}$, $\theta$, $\eta$, $t$ and $\nabla_{\theta}$ represents loss, model parameters, learning rate, iteration step and gradient update, respectively. This aggregation smoothens the overall label prediction and enhances the quality of the pseudo labels for the target domain. 

Temporal ensembling is a robust mechanism for identifying and learning domain-invariant features, thereby reducing discrepancies between the source and target data over time. The model's adaptation to the target domain improves significantly by averaging predictions over time. Temporal ensembling mitigates class imbalance~\cite{de2023systematic} by generating pseudo-labels and enabling stable learning, which helps in learning a balanced data representation and alleviates the skew toward majority classes seen in the labeled data. We determine the optimal setting for \textbf{\textit{$\alpha$}} using a grid-search method and then fine-tuned it for the optimal performance across the datasets.

\vspace{-0.3cm}
 \begin{equation}
 \theta_{t+1} = \theta_t - \eta \cdot \nabla_{\theta}\mathcal{L}
 \label{eq:weight_update}
 \end{equation}


\subsection{UDA via kernel-based class-wise conditional mean maximum discrepancy~(kCMMD)} 


We focus on aligning conditional distributions using our proposed kernel-based class-wise conditional mean maximum discrepancy (kCMMD) approach in the Reproducing Kernel Hilbert Space (RKHS)~\cite{ren2016conditional}. In RKHS, conditional distributions are mapped to a high-dimensional feature space, and $k$CMMD computes the mean discrepancy between these embeddings to minimize domain shifts. This reduces the conditional probability gap between the source ($P^{s}(Z|Y)$) and target ($P^{t}(Z|Y)$) domains, establishing a robust conditional domain-invariant feature space. The generalized $k$CMMD loss is defined in Eq.~\ref{eq:kcmmd_gen}, where $C$ represents the number of classes, and $n_c$ and $m_c$ denote the number of samples from class $c$ in the source and target domains, respectively, with $x_i^s$, $x_j^s$, $x_i^t$, and $x_j^t$ representing feature embeddings.

\begin{equation}
\begin{aligned}
\mathcal{L}_{\text{kCMMD}} = \sum_{c=1}^{C} \Bigg[ \frac{1}{n_c(n_c - 1)} \sum_{\substack{i, j = 1 \\ i \ne j}}^{n_c} K(\phi)(x_i^s, x_j^s) + \frac{1}{m_c(m_c - 1)} \\ \sum_{\substack{i, j = 1 \\ i \ne j}}^{m_c} K(\phi)(x_i^t, x_j^t) - \frac{2}{n_c m_c} \sum_{i=1}^{n_c} \sum_{j=1}^{m_c} K(\phi)(x_i^s, x_j^t) \Bigg]
\label{eq:kcmmd_gen}
\end{aligned}
\end{equation}




Considering our problem scenario, where the source domain is represented by a feature set $x_s \in \mathbb{R}^{n \times d}$ with corresponding labels $y_s$, and a target domain represented by $x_t \in \mathbb{R}^{m \times d}$ with pseudo-labels $y_{\Bar{t}}$, the $k$CMMD loss focuses on minimizing distribution differences for each class $c \in C$. $k$CMMD utilizes the Radial Basis Function (RBF) kernel, defined as $K(\phi)(x, x') = \exp(-\gamma |x - x'|^2)$, where $\gamma$ is the kernel bandwidth parameter. For each class $c$ from the source~($X_s$) and target~($X_t$) domains, we compute the following kernel matrices ($K_M$), quantifying intra- and inter-domain discrepancies using element-wise multiplication~[$\otimes$]:
\vspace{-0.2cm}
\begin{align*}
\text{\textbf{Source-Source}} [K_M^{ss}] & = K(\phi)(x, x') \otimes(X_s^{(c)}, X_s^{(c)}, \gamma) + \lambda I \\
\text{\textbf{Target-Target}} [K_M^{tt}] & = K(\phi)(x, x') \otimes(X_t^{(c)}, X_t^{(c)}, \gamma) + \lambda I \\
\text{\textbf{Source-Target}} [K_M^{st}] & = K(\phi)(x, x') \otimes(X_s^{(c)}, X_t^{(c)})
\end{align*}

Here, $\lambda I$ denotes the regularization constant added to the diagonals of the source-source and target-target kernel matrices to prevent issues like ill-conditioning, control overfitting~\cite{ren2016conditional, ren2019learning} and class-imbalance during intra-domain computation. This regularization ensures that the learned features are adaptable and generalizable across users. Kernel methods are instrumental in capturing non-linear features and mapping data into higher-dimensional spaces so non-linear relationships can be linearly separated. The $k$CMMD loss is calculated as the mean of discrepancies across all classes using a function $\delta$, which computes the mean discrepancy for each class. The overall regularized $k$CMMD discrepancy loss function for the kernel matrices $K_M^{ss}$, $K_M^{tt}$, and $K_M^{st}$ is defined in Equation~\eqref{eq:cmmd}:
\vspace{-0.2cm}
\begin{equation}
\mathcal{L}_{kCMMD} = (\Bar{K_M^{ss}} + \Bar{K_M^{tt}} - 2 \cdot \Bar{K_M^{st}})
\label{eq:cmmd}
\end{equation}

Equation~\ref{eq:cmmd} measures the overall conditional discrepancy between the source and target domain distributions. Regularized $K_M^{ss}$ and $K_M^{tt}$ represent the mean intra-domain similarities for the source and target domains, respectively, while $K_M^{st}$ quantifies the mean inter-domain similarity. The equation calculates domain discrepancy by adding the mean similarities within each domain and subtracting twice the mean cross-domain similarity, i.e. ensuring a bidirectional symmetrical step to seek to minimize differences between the source and target domains without bias. When the source and target domain distributions align closely, the values of $K_M^{ss}$ and $K_M^{tt}$ will approximate $K_M^{st}$, resulting in a lower domain discrepancy value for effective cross-user adaptation task.

\subsection{Unsupervised Data Augmentation Module}~\label{sec:uda_aug}


We address user variability such as differences in activity execution, proficiency levels and individual traits by integrating an unsupervised data augmentation approach~\cite{xie2020unsupervised} to build robustness against cross-user variations. Our augmentation module ensures that (i) augmented samples are valid and realistic, (ii) they exhibit diversity, and (iii) they introduce specific inductive biases. Recognizing that sudden temporal shifts in wearable data can impair model performance, we develop a model invariant with such transformations. To achieve this, we implement two geometric-based augmentations: \textbf{\textit{jitter}} introduces high-frequency fluctuations in the data, distinct from actual motion, thereby adding stochastic noise and \textbf{\textit{rotation}} which involves angular displacement or orientation change, quantifying the degree and axis of rotational movement which are well-established for capturing real-world variations in wHAR tasks~\cite{um2017data}. Let $G$ denote the geometric augmentation applied to the source and target data, $x_s$ and $x_t$, defined as $G = [r_{\theta}, \sigma]$. The augmented samples are then $x_s' = G \cdot x_s$ and $x_t' = G \cdot x_t$. Our goal is to enforce consistency between the model's predictions on original and augmented data, ensuring robustness to user-specific variations. We employ a consistency loss (Eq.~\eqref{eq:consistency}), where $f_{\theta}(x_s)$, $f_{\theta}(x_s')$, $f_{\theta}(x_t)$, and $f_{\theta}(x_t')$ represent predictions for real and augmented samples. This approach aligns with the Lipschitz continuity theorem~\cite{gouk2021regularisation}, ensuring the model's outputs change proportionally with input variations, enhancing generalization to diverse user variations.

\vspace{-0.2cm}
\begin{equation}
\begin{aligned}
\mathcal{L}_{\text{C}} = \mathbb{E}{x_s \sim D_s, x'_s \sim t(x_s)} \left[ D_{KL}(f_{\theta}(x_s) || f_{\theta}(x'_s)) \right] + \\ 
\mathbb{E}{x_t \sim D_t, x'_t \sim t(x_t)} \left[ D_{KL}(f_{\theta}(x_t) || f_{\theta}(x'_t)) \right]\label{eq:consistency}
\end{aligned}
\end{equation}

\subsection{\name Training Procedure }

The overall aim is to obtain a set of optimized model parameters trained on labeled source data that work well on the target domain and generate high-quality pseudo labels for the target domain. To generate high-quality pseudo labels, we utilize the ensembling approach due to the ability to smoothen noisy pseudo labels. Additionally, we adopted a conditional probability alignment approach, which provides a robust domain-invariant feature embedding. We highlight the overall joint optimization objective function for \name framework enumerated in Equation~\eqref{eq:overall}. In the overall formulation, $\theta$ encompasses all the parameters within the deep network. The hyperparameters $\beta_{0}$ and $\beta_{1}$ are set to 1, facilitating the fine-tuning of the overall objective function.

\vspace{-0.2cm}
\begin{equation}
 \min_{F_{\theta}} \mathcal{L}_{overall}  =  \underbrace{\mathcal{L}_{SL} + \beta_0 \mathcal{L}_{kCMMD}}_\text{Domain Adaptation} \ \ + \underbrace{\beta_1\mathcal{L}_{C}}_\text{Consistency Regularizer}
 \label{eq:overall}
\end{equation}


\vspace{-0.5cm}
\section{Experiments} \label{sec:experiment}

We evaluate \names~via extensive experiments on the problem of limited supervision settings for ADLs action recognition by investigating the following research questions (RQs): (i) \textbf{RQ1 (Accuracy):} How well is \name performing on publically available datasets? (ii) \textbf{RQ2 (Robustness):} How can we adapt \name for accurate recognition across different users' proficiency levels in wHAR? (iii) \textbf{RQ3 (Compatibility):} Is \name optimized for high quality pseudo-labels generation? (iv) \textbf{RQ4 (Comparative Analysis):} Is \name minimizing conditional data distribution discrepancies?



\begin{table}
\small
\centering
\caption{List of \textit{$\mu$DAR's} Hyperparameters}
\label{tab:hyperparameters}
\scalebox{0.82}{
\begin{tabular}{lr} 
\toprule
\textbf{Hyper-parameter} & \textbf{Values} \\ 
\midrule
No. of maximum convolution layers & 1,2,3 \\ 
No. of filters in convolution layers & 32, 128, 64 \\ 
Convolution filter dimension & 5x1,5x1,5x1 \\ 
No. of maximum fully connected layers & 2 \\ 
No. of neurons in fully connected layers & 8, 4 \\ 
Batch size & 64 \\ 
Dropout rate & 0.3 \\ 
Optimizer & Adam \\ 
Learning Rate & 0.001 - 0.0003 \\ 
Max number of epochs & 128 \\ 
$\lambda$ value & 0.18 - 0.45 \\ 
$\alpha$ value & 0.55 - 0.75 \\ 
$\sigma$ value & 0.01 - 0.10 \\
$r_{\theta}$ value &  5$^{\circ}$ - 45$^{\circ}$ \\
\bottomrule
\end{tabular}}
\end{table}

\begin{table*}
\centering
\caption{Comparison of $\mu$DAR's Macro F1 Score ($\pm$ SD) and Average Improvement Margin vs. SWL-Adapt~\cite{hu2023swl}, ContrasGAN~\cite{sanabria2021contrasgan}, CoTMix~\cite{eldele2022cotmix},  HoMM~\cite{chen2020homm}, DUA~\cite{mirza2022norm} and AdvSKM~\cite{liu2021adversarial} across BAR, DSADS, PAMAP2, and MMDOS datasets. The top and 2nd best results are in bold and underlined, respectively.}
\label{tab:public_results}
\resizebox{.995\textwidth}{!}{%
\begin{tabular}{@{}lcccccccc@{}}
\toprule
Datasets &
  SWL-Adapt~\cite{hu2023swl} &
  ContrasGAN~\cite{sanabria2021contrasgan} &
  CoTMix~\cite{eldele2022cotmix} & AdvSKM~\cite{liu2021adversarial} & HoMM~\cite{chen2020homm}& DUA~\cite{mirza2022norm}  &
  \textbf{\name (Our)} &
  \textbf{Average Improvement Margin} \\ \midrule
\textbf{DSADS~\cite{altun2010human}}  & {0.70 $\pm$ (0.0188)} &  0.64 $\pm$ (0.0213)             & 0.61 $\pm$ (0.0252) & {0.71 $\pm$ (0.0132)} & {0.68 $\pm$ (0.0265)}& \underline{0.73 $\pm$ (0.0105)} & \textbf{0.77 $\pm$ (0.0239)} & {0.048 $\pm$ (0.015)} \\
\textbf{PAMAP2~\cite{reiss2012creating}} & \underline{0.73 $\pm$ (0.0174)} & 0.72 $\pm$ (0.0160)    & 0.66 $\pm$ (0.0248) & 0.70 $\pm$ (0.0213) & {0.67 $\pm$ (0.0229)} & {0.72 $\pm$ (0.0095)} & \textbf{0.78 $\pm$ (0.0211)} & {0.054 $\pm$ (0.011)} \\
\textbf{MMDOS~\cite{sharshar2022mm}}  & 0.72 $\pm$ (0.0181)             & \underline{0.78 $\pm$ (0.0159)} & 0.68 $\pm$ (0.0227) & 0.69 $\pm$ (0.0193) & {0.70 $\pm$ (0.0162)} & {0.76 $\pm$ (0.0122)} & \textbf{0.84 $\pm$ (0.0256)} & {0.061 $\pm$ (0.013)} \\ 
\textbf{BAR~\cite{ghosh2022decoach}}  & 0.68 $\pm$  (0.0219)       & {0.71 $\pm$ (0.0192)} & \underline{0.73 $\pm$ (0.0205)} & {0.70 $\pm$ (0.0125)}  & {0.65 $\pm$ (0.0217)} & {0.69 $\pm$ (0.0155)}   &  \textbf{0.82 $\pm$ ((0.0198)} & {0.097 $\pm$ (0.028)} \\ 
\bottomrule
\end{tabular}}
\end{table*}

\subsection{Setup}
\subsubsection{Datasets} We selected \textbf{four} publicly available datasets (1) {\bf Badminton Activity Recognition~(BAR) Dataset~\cite{ghosh2022decoach}}, capturing 12 badminton strokes from 11 subjects in controlled and uncontrolled environments, with 30 iterations per stroke. (2) {\bf Daily and Sports Activity~(DSADS) Dataset~\cite{altun2010human}}, featuring 9-DOF sensor data from eight individuals performing 19 activities, recorded at 25 Hz. (3) {\bf Physical Activity Monitoring~(PAMAP2) Dataset~\cite{reiss2012creating}}, consisting of 18 activities from nine subjects, captured at 100 Hz from three sensors. (4) {\bf Workout Activities~(MM-DOS) Dataset~\cite{sharshar2022mm}}, with IMU data from 50 participants performing four gym workouts, recorded at 50 Hz across nine body locations.


\subsubsection{Baselines} \label{sec:baseline} We evaluate and contrast the performance of \name with \textbf{six} SOTA UDA algorithms including \textbf{\textit{SWL-Adapt}}~\cite{hu2023swl}, \textbf{\textit{ContrasGAN}}~\cite{sanabria2021contrasgan}, \textbf{\textit{AdvSVM}}~\cite{liu2021adversarial}, \textbf{\textit{HoMM}}~\cite{chen2020homm}, \textbf{\textit{DUA}}~\cite{mirza2022norm} and \textbf{\textit{CoTMix}}~\cite{eldele2022cotmix}. 

\subsection{Implementation Details}

\subsubsection{Data Pre-Processing} 

This study uses raw signals from accelerometers, gyroscopes, and magnetometers in a body-worn IMU sensor network as input features. To address motion artifacts, we applied a median filter and normalized 48 features using a min-max scaler for the {\it BAR} dataset, which includes data from three sensor axes. Due to signal clipping beyond $\pm 2g$ in the low-noise accelerometer, we used wide/high-range accelerometers to capture rapid movements more effectively. For feature extraction, we applied a sliding window technique with varying overlap and sampling rates across the {\it BAR}, {\it PAMAP2}, {\it DSADS}, and {\it MMDOS} datasets to remove artifacts and extract temporal patterns. Labeling was done using majority voting within each window. We adopted a grid search fine-tuned window sizes and overlaps, ranging from 5\% to 95\% overlap and window durations from 0.1 to 1 second, to optimize parameter selection for capturing complex patterns across datasets.


\subsubsection{Network Architecture and Training}

Our experiments were conducted on a Linux server with an Intel i7-6850K CPU, 4x NVIDIA GeForce GTX 1080Ti GPUs, and 64 GB RAM. We used Python, specifically PyTorch, for data preprocessing and deep learning tasks. Given the class imbalance in most datasets, we adopted the macro-F1 score as the primary evaluation metric. The datasets were split into training, validation, and testing sets (60-20-20\%), and hyperparameters were optimized using the validation set. Notably, the validation and test sets were not utilized during training.

\subsection{Results and Discussion}
\subsubsection{\textbf{Classification Performance (RQ1)}}
In this analysis, we compare and contrast \textit{$\mu$DAR} against six state-of-the-art (SOTA) UDA algorithms, demonstrating a significant average improvement of \textbf{$\approx$ 4-12\%} across four public datasets. Our results distinguish the top-performing and second-best methods, revealing that SWL-Adapt frequently surpasses other UDA algorithms, likely due to its meta-optimized loss function, which assigns sample weights, enhances domain alignment, and improves cross-user adaptation, as shown in Table~\ref{tab:public_results}.



Notably, \textit{\name} excels across all datasets, covering activities from complex gym and sports-related actions to ADLs and iADLs. Its strength lies in using temporal ensembling to generate reliable pseudo-labels and integrating kernel-based class-wise conditional mean maximum discrepancy (kCMMD) to enhance domain adaptation. Additionally, performance improves with consistency regularization, effectively handling augmented sample variations in real-world settings and further supporting analyses are presented in section~\ref{sec:comparative}.

\begin{figure}[H]
\begin{center}
\begin{subfigure}[c]{0.60\linewidth}
    \includegraphics[scale=0.167]{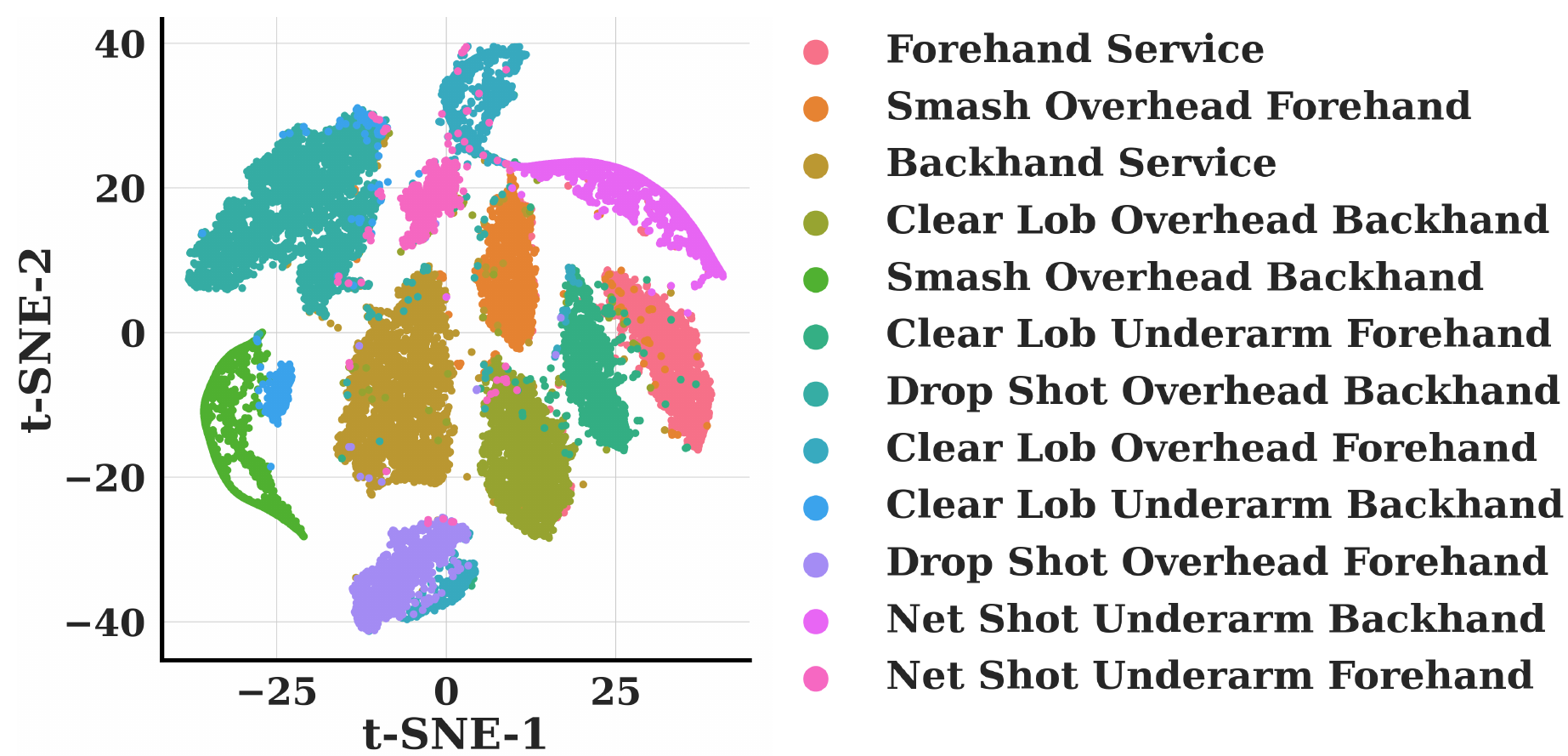}
    \caption{Feature Embedding on Target User using~\textit{$\mu$DAR} on the BAR dataset.}
    \label{fig:feature_repre}
\end{subfigure}%
 \hfill
 \begin{subfigure}[c]{0.38\linewidth}
   \includegraphics[scale =0.128]{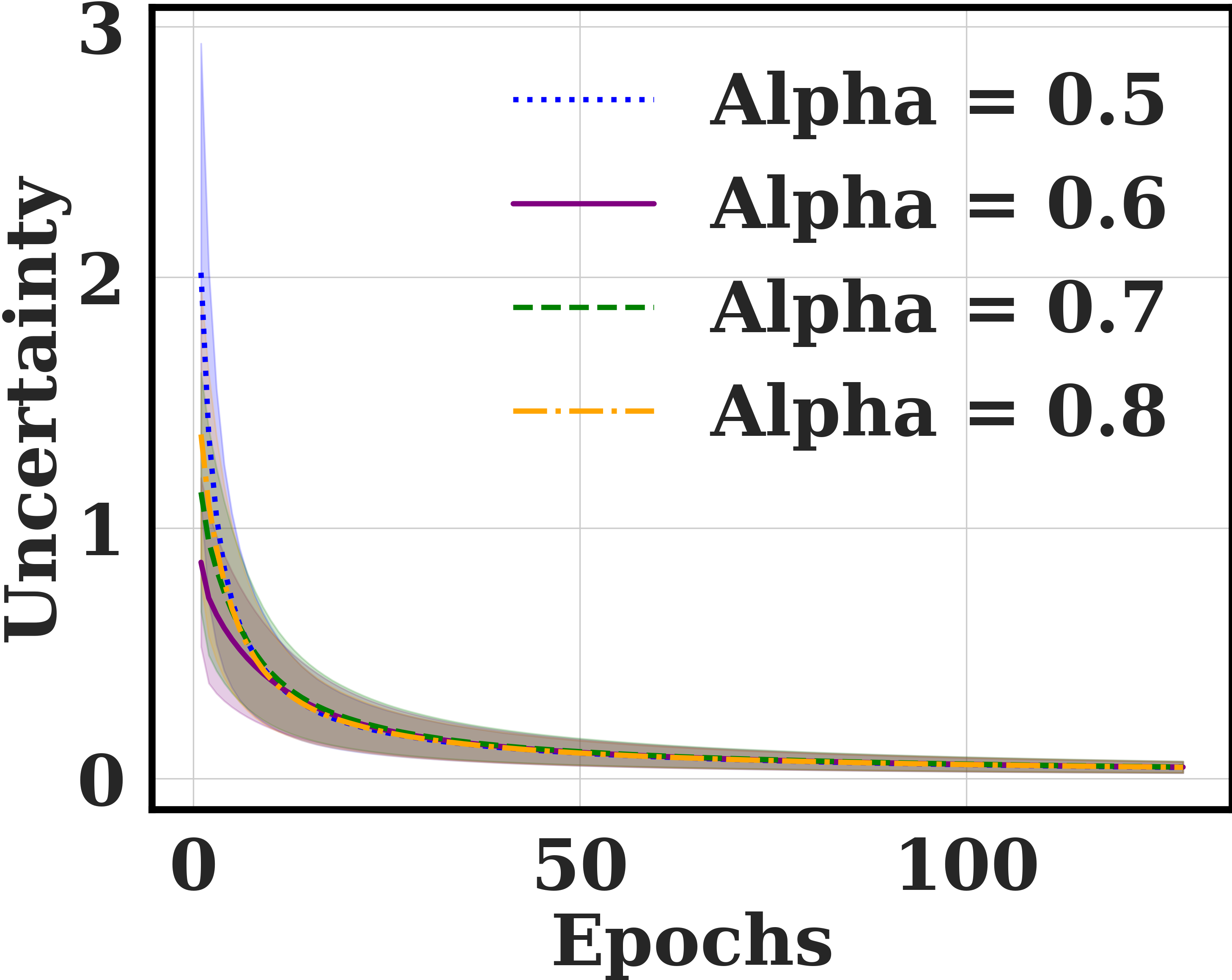}
    \caption{Correlation between $\alpha$ parameter and uncertainty.}
    \label{fig:uncertaintiy}
 \end{subfigure}%
 \caption{Ablation Analysis for \textbf{RQ1} and \textbf{RQ3}, respectively}
\end{center}
\end{figure}

\subsubsection{\textbf{Robustness of \name to User-Variations (Proficiency levels) (RQ2)}}

We assess the robustness and scalability of the \name framework by comparing across SOTA baseline algorithms for handling cross-user proficiency variations. The experiment trains models using expert data (source users) and evaluates their performance on beginner data (target users). Results show that \textit{$\mu$DAR} significantly outperforms benchmarks, with an average macro F1 score improvement of {\bf [7.3 $\pm$ (1.54)]\%}. \textit{$\mu$DAR's} success is due to its effective use of temporal ensembling, where the $\alpha$ parameter helps identify unique instance-based features across genders. By incorporating $k$CMMD loss and consistency, regularization captures non-linear features in high-dimensional space and ensures consistent predictions. Further analysis on optimizing the $\alpha$ parameter and its impact on pseudo-label quality is highlighted in section~\ref{sec:optimze_alpha}.


\begin{figure}
\begin{center}
\begin{minipage}[c]{0.32\linewidth}
 \includegraphics[scale=0.12]{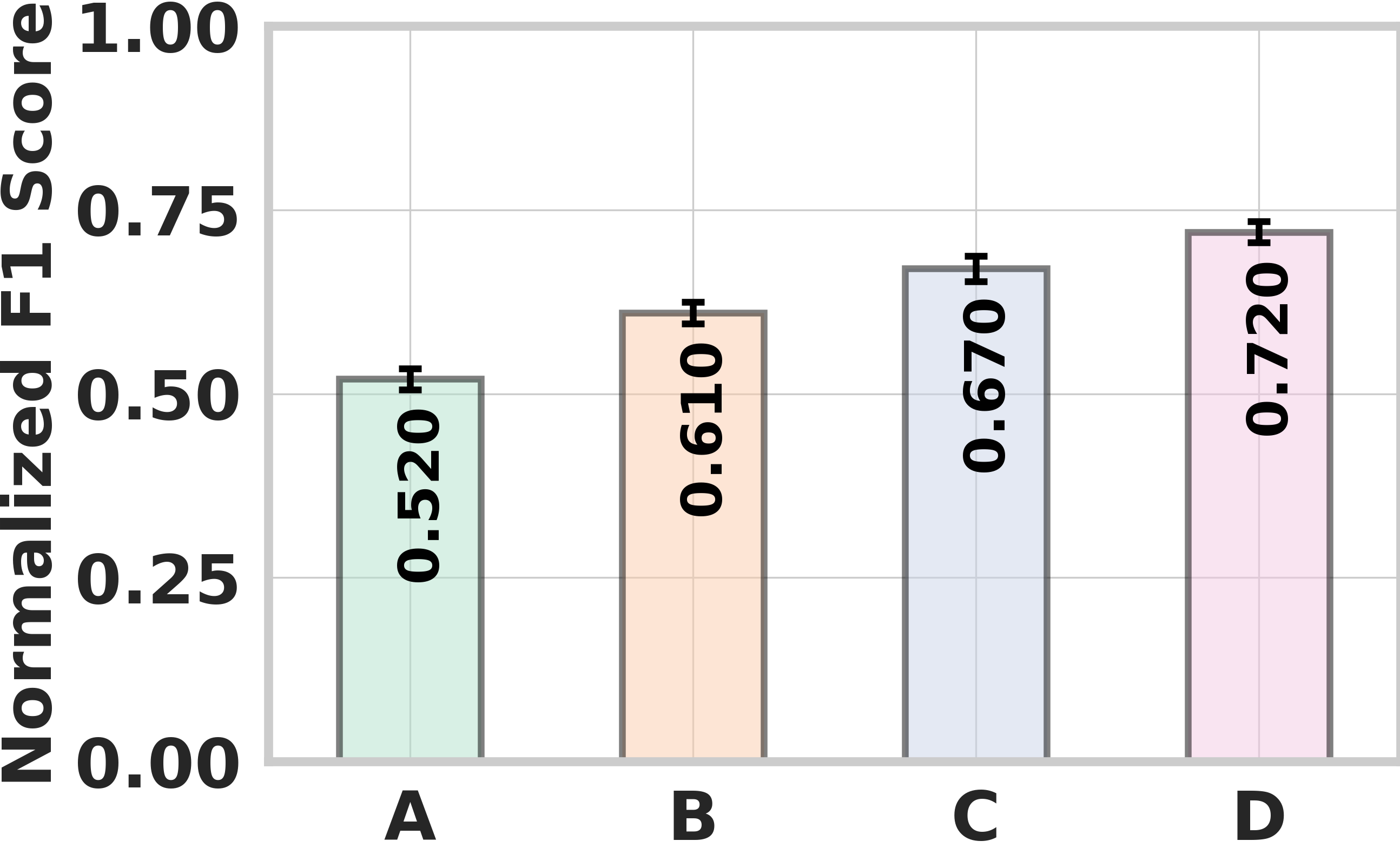}
\subcaption{Augmentation on target domain}
\label{fig:t_da}
    \end{minipage}
\hfill
\begin{minipage}[c]{0.32\linewidth}
 \includegraphics[scale=0.12]{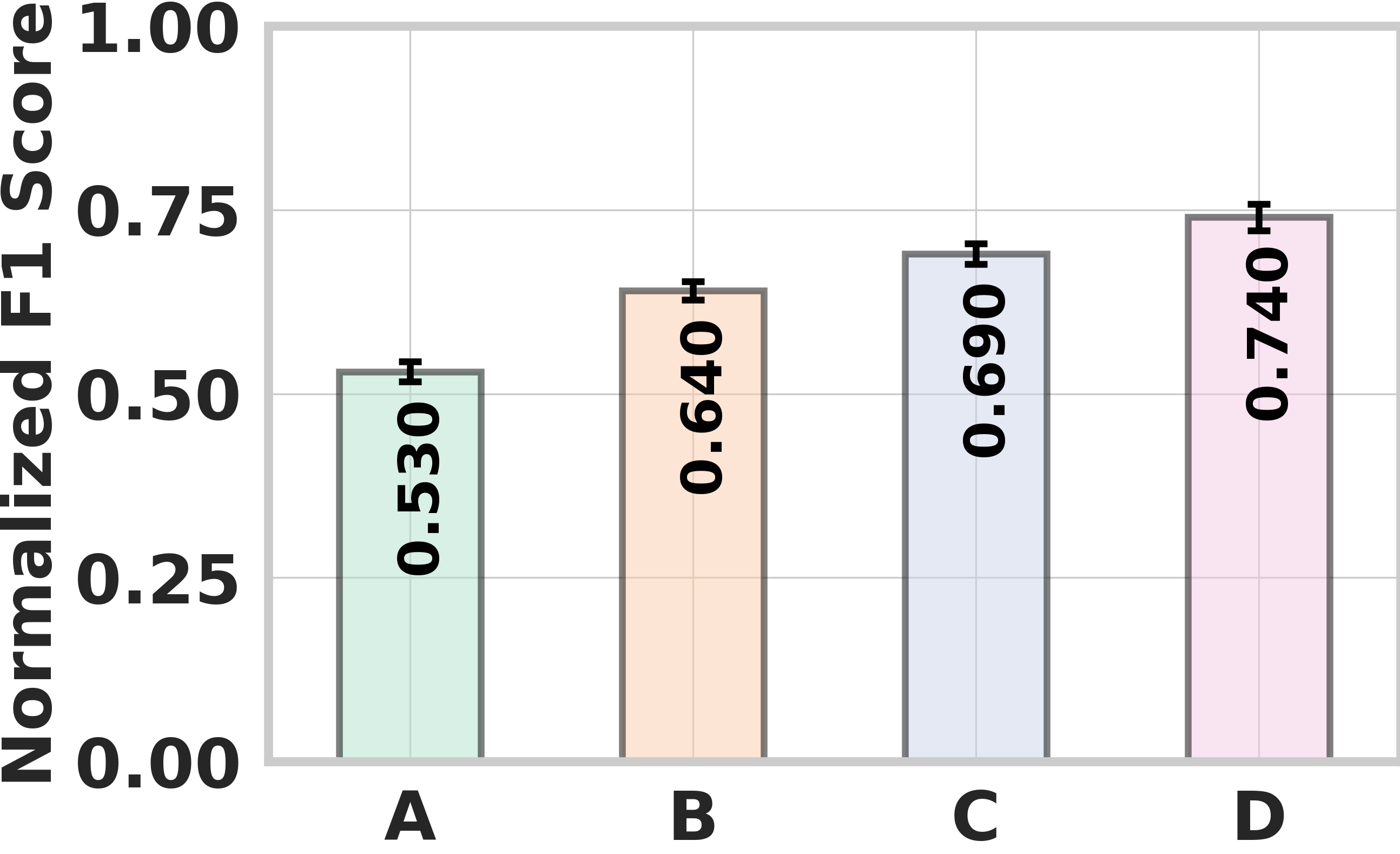}
\subcaption{Augmentation on source domain}
\label{fig:s_da}
\end{minipage}%
\hfill
\begin{minipage}[c]{0.32\linewidth}
 \includegraphics[scale=0.12]{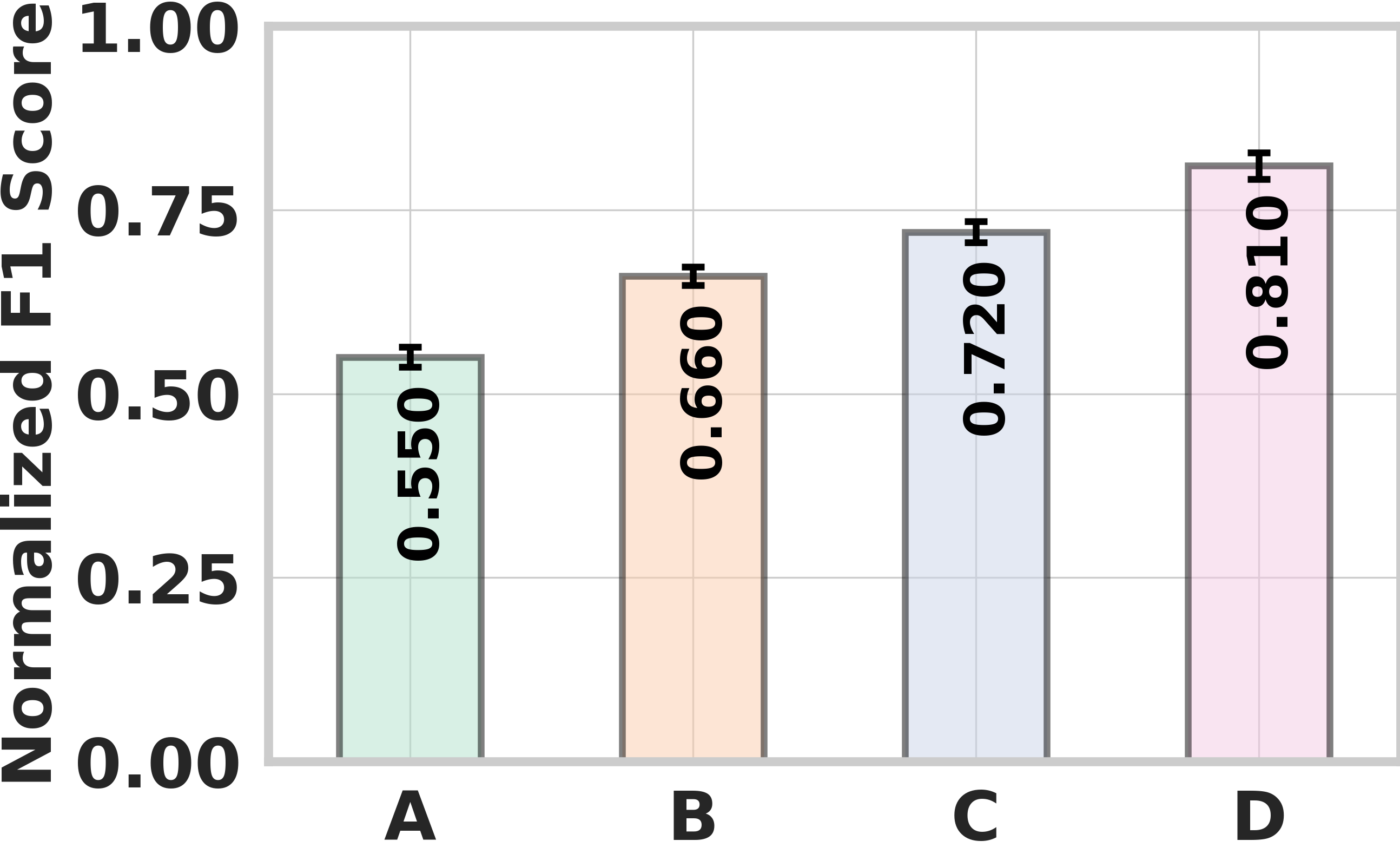}
\subcaption{Augmentation on source and target domains}
\label{fig:s_t_da}
\end{minipage}%
\caption{The ablation study highlighting the performance difference between \textbf{(A), (B), (C), (D)} corresponds to \textit{Baseline}, \textit{{Temporal Ensembling}}, \textit{Temporal Ensembling + $k$CMMD loss }and \textit{Temporal Ensembling + $k$CMMD loss + Consistency Regularizer}, respectively across all the benchmark datasets.}
\label{fig:ablation_temp}
\end{center}
\end{figure}

We present the cross-proficiency user adaptation performance comparing the best accuracy SOTA SWL-Adapt algorithm and \name using the MM-DOS dataset and \name achieves an approximate {\bf 11\%} improvement over SWL-Adapt as shown in Figure~\ref{fig:mmdos_ablation_cf}. Additionally, the results demonstrate that \name outperforms SOTA UDA algorithms in the wHAR domain across cross-user and cross-dexterity variations.

\begin{figure}
\begin{center}
\begin{subfigure}[c]{0.48\linewidth}
    \includegraphics[scale=0.106]{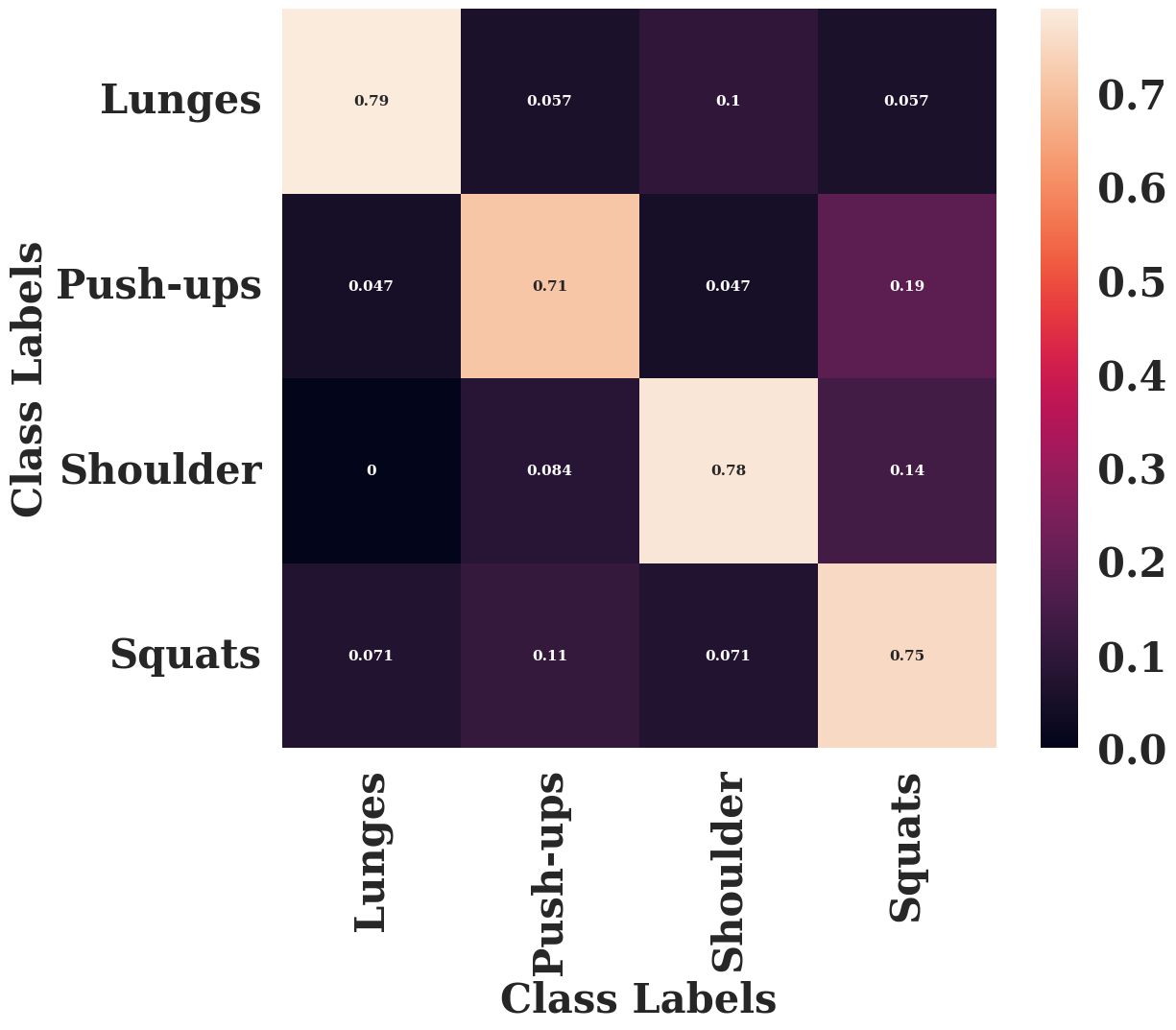}
    \caption{SWL-Adapt~\cite{hu2023swl}}
     \label{fig:kcmmd_temp_auroc_mmdos}
 \end{subfigure}%
 \hfill
 \begin{subfigure}[c]{0.48\linewidth}
   \includegraphics[scale =0.153]{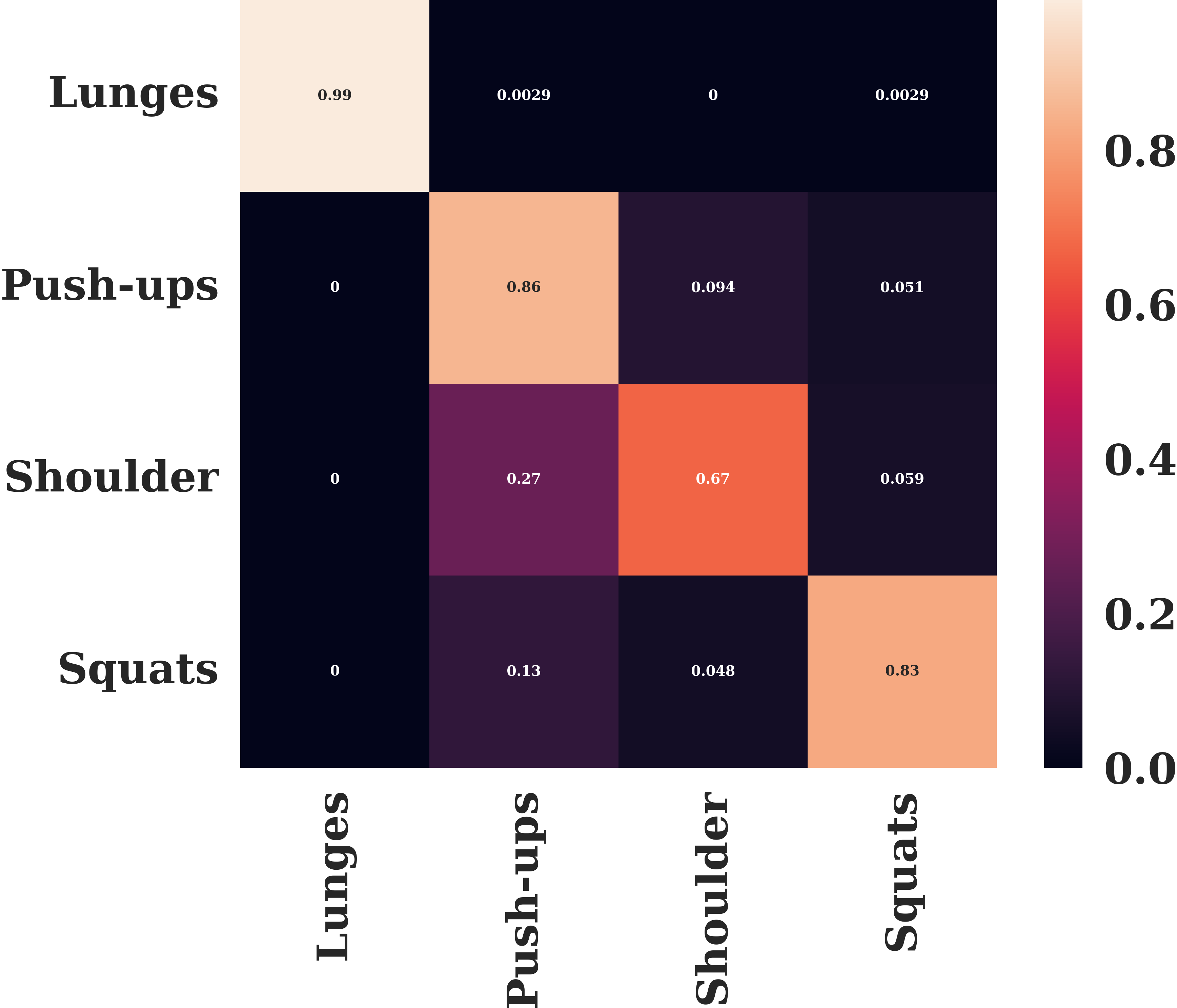}
    \caption{\name (Our)}
     \label{fig:kcmmd_temp_kcmmd_auroc_mmdos}
 \end{subfigure}%
 \caption{Confusion matrices for cross-dexterity levels (expert~(source) $\rightarrow$ beginner~(target)) on MMDOS dataset.}
 \label{fig:mmdos_ablation_cf}
\end{center}
\end{figure}

\subsubsection{\textbf{Compatibility Study (RQ3)}}\label{sec:optimze_alpha}

In this study, we evaluated the effectiveness of self-ensembling for generating reliable pseudo-labels in the target domain. We used \textbf{entropy}, \{$H(x) = -\sum_{i=1}^{n} p(x_i) \log(p(x_i))$\}, to quantify prediction uncertainty. As shown in Figure~\ref{fig:uncertaintiy}, there is an inverse relationship between the $\alpha$ parameter of ensembling and entropy: higher $\alpha$ leads to lower uncertainty, indicating more stable pseudo-labels. Our experiments found that an $\alpha$ range of \textbf{\textit{[0.55-0.75]}} is optimal for generating high-quality pseudo-labels. Figure~\ref{fig:uncertaintiy} further validates that an $\alpha$ of \textbf{0.60} achieves faster convergence, emphasizing the critical role of $\alpha$ in improving pseudo-label quality for the target domain.

\subsubsection{\textbf{Comparative Analysis (RQ4)}}\label{sec:comparative}

This analysis evaluates the individual and combined effects of temporal ensembling, $k$CMMD loss, and consistency regularization on~\textit{$\mu$DAR}'s performance across four SOTA datasets. We report normalized F1 scores to demonstrate the synergy across three augmentation settings, as shown in Figure~\ref{fig:ablation_temp}, highlighting the generalizability and robustness of \textit{$\mu$DAR}. Our results show a significant improvement of \textbf{[22-26\%]} when incorporating all methods over the baseline (no adaptation), as depicted in Fig.~\ref{fig:t_da},~\ref{fig:s_da}, and~\ref{fig:s_t_da}. This synergy enhances overall performance, advancing domain generalization for the wHAR domain.

\vspace{-0.2cm}
\section{Conclusion and Future Work}\label{sec:conclusion}
This work introduces a novel single-stage joint optimization strategy for cross-user wHAR domain. Our approach integrates \textit{temporal-ensembling} with kernel-based class-wise conditional mean maximum discrepancy (\textit{$k$CMMD}), effectively generating high-quality pseudo labels and minimizing conditional domain discrepancies. Additionally, using KL divergence for consistency regularization stabilizes label predictions in the presence of augmented samples. The \name framework demonstrated notable superiority over benchmark UDA algorithms in extensive experiments, showing generalizability and robustness, outperforming SOTA approaches by \textbf{$\approx$ 4-12\%}.

In the future, we plan to expand the capabilities of the \name framework by exploring higher-order statistics-based methods, such as cross-covariance estimation for conditional embedding alignment. These enhancements will advance domain generalization in the wHAR domain, paving the way for more versatile and effective action recognition systems.



\vspace{-0.2cm}
{
\scriptsize
\bibliographystyle{plainnat}
\bibliography{reference.bib}
}

\end{document}